\documentclass{article}

\usepackage{PRIMEarxiv}

\usepackage[utf8]{inputenc} 
\usepackage[T1]{fontenc}    
\usepackage{hyperref}       
\usepackage{url}            
\usepackage{booktabs}       
\usepackage{amsfonts}       
\usepackage{nicefrac}       
\usepackage{microtype}      
\usepackage{lipsum}
\usepackage{fancyhdr}       
\usepackage{graphicx}       
\graphicspath{{media/}}     

\usepackage{amsmath}
\usepackage{amsfonts}
\usepackage{amsthm}
\usepackage{float}
\usepackage{geometry}
\usepackage[numbers]{natbib}
\usepackage{subcaption}
\usepackage{multirow}
\usepackage{booktabs}

\title{SatSplatYOLO: 3D Gaussian Splatting-based Virtual Object Detection Ensembles for Satellite Feature Recognition
}

\author{
  Van Minh Nguyen \\
  NEural TransmissionS (NETS) Lab \\
  Florida Institute of Technology \\
  \texttt{vnguyen2014@my.fit.edu} \\
   \And
  Emma Sandidge \\
  NEural TransmissionS (NETS) Lab \\
  Florida Institute of Technology \\
  \texttt{esandidge2020@my.fit.edu} \\
   \And
  Trupti Mahendrakar \\
  NEural TransmissionS (NETS) Lab \\
  Florida Institute of Technology \\
  \texttt{tmahendrakar2020@my.fit.edu} \\
   \And
  Ryan T. White \\
  NEural TransmissionS (NETS) Lab \\
  Florida Institute of Technology \\
  \texttt{rwhite@fit.edu} \\
}

\begin{document}
\maketitle

\begin{abstract}
On-orbit servicing (OOS), inspection of spacecraft, and active debris removal (ADR). Such missions require precise rendezvous and proximity operations in the vicinity of non-cooperative, possibly unknown, resident space objects. Safety concerns with manned missions and lag times with ground-based control necessitate complete autonomy. In this article, we present an approach for mapping geometries and high-confidence detection of components of unknown, non-cooperative satellites on orbit. We implement accelerated 3D Gaussian splatting to learn a 3D representation of the satellite, render virtual views of the target, and ensemble the YOLOv5 object detector over the virtual views, resulting in reliable, accurate, and precise satellite component detections. The full pipeline capable of running on-board and stand to enable downstream machine intelligence tasks necessary for autonomous guidance, navigation, and control tasks.
\end{abstract}

\keywords{Space Domain Awareness \and Satellite inspection \and 3D Gaussian Splatting \and You Only Look Once \and 3D scene rendering \and Object detection}

\section{Introduction}

The increasing amount of space debris and its threats to the spacecraft in orbit, the problem of performing autonomous rendezvous and proximity operations (RPO) around unknown non-cooperative spacecraft to prevent the formation of more debris has garnered great interest. To tackle this problem, there is a pressing need for on-orbit servicing (OOS) and active debris removal (ADR) to capture and de-orbit or extend satellite lifetimes as the bulk of space debris consists of large defunct satellites, which may be fragile or unstable. The complexity of the situation and lag times associated with ground-based control necessitates complete autonomy during the rendezvous and proximity operation (RPO) phases. This requires robust characterization of the geometry of non-cooperative resident space objects (RSOs) to enable downstream computer vision and machine intelligence tasks necessary for autonomous guidance, navigation, and control tasks.

This article aims to characterize the geometry and recognize features of a satellite on orbit. This will be accomplished in three stages: (1) use data captured along an orbit to learn a 3D scene of the satellite using 3D Gaussian Splatting (3DGS), (2) render synthetic images from multiple novel viewing angles perturbed from real views, and (3) ensemble object detections across the renders using YOLOv5 to produce high-quality satellite feature classification and localization. The end-to-end computational footprint is low enough to run on low-powered spaceflight hardware.


The primary contributions of this article include:
\begin{enumerate}
    \item Systematic camera generation for 3DGS rendering of satellites for full 3D characterization
    \item A rendering-based ensembling technique for high-quality satellite component detection
    \item Hardware-in-the-loop experiments with realistic lighting and motion conditions for known and unknown satellites
\end{enumerate} 

\section{Related Work}
\label{sec:rw}

\subsection{Object Detection}


Object detection is a computer vision problem that aims to localize and classify objects from a predefined set of classes. The input to an object detector is an image and the output is a set of bounding boxes tightly surrounding the objects along with object class probabilities and confidences.

Neural network-based object detectors were introduced in 2013 \cite{szegedy_deep_2013, sermanet_overfeat_2014} and have achieved state-of-the-art object detection performance on all major benchmark datasets \cite{noauthor_papers_nodate, noauthor_papers_nodate-1}. Architectural advancements \cite{he_spatial_2015, he_deep_2016, lin_feature_2017, liu_path_2018, wang_cspnet_2020} and GPU acceleration \cite{krizhevsky_imagenet_2012} have led to more efficient single-stage detectors \cite{liu_ssd_2016, redmon_you_2016} that can run in low-compute environments, e.g. on spaceflight hardware. The most effective of these are from the You Only Look Once (YOLO) family \cite{redmon_you_2016}. YOLO's successors have set most current state-of-the-art real-time object detection marks in different domains, with all of the following being competitive: YOLOv4 \cite{bochkovskiy_yolov4_2020}, YOLOv5 \cite{jocher_yolov5_2020}, YOLOv6 \cite{li_yolov6_2022}, YOLOv7 \cite{wang_yolov7_2023}, and YOLOv8 \cite{jocher_yolov8_2023}. There is not a universal leader at the present time.




\subsection{3D Scene Rendering}

\textbf{Neural Radiance Fields (NeRFs)} NeRFs \cite{mildenhall_nerf_2021} employ a fully connected neural network to learn a radiance field that determines saturation and colors within a scene. The input is a 5D coordinate that specifies the location, $(x,y,z)$ and viewing angles, $(\theta , \phi )$ of a scene. The model outputs color and density, and rays are traced through the scene to render a 2D image of the 3D scene representation using volume rendering techniques. NeRFs are differentiable end-to-end, allowing them to be trained with gradient-based methods. While NeRFs are effective at generating high-quality renders, they have slow training and rendering times.

\textbf{Instant NeRF} Instant NeRF \cite{muller_instant_2022} improves NeRF training by accelerating the training process. During the input process, the model learns a multi-resolution set of feature vectors, so the data is encoded faster. This results in lower training times and a smaller neural network. This means that Instant NeRF can produce similar outputs to NeRF with a lower computational cost in training, however it still has rendering cost similar to regular NeRF.

\textbf{3D Gaussian Splatting (3DGS)} 3DGS \cite{kerbl_3d_2023} is a newer method that is able to learn a 3D representation of a scene using point-based Gaussians. It trains at a slightly faster rate than Instant NeRF but it is able to render scenes at a much lower computational cost than the previous methods. The algorithm uses Structure-from-Motion (SfM) \cite{schonberger_structure--motion_2016} to build a 3D point-cloud representation of a scene based on given input images. These points are then represented with 3D Gaussian random variables that are trained to represent the scene. This reduces training times as there is no longer a large neural network with many layers to train. In addition, the point-based rendering of 3DGS avoids NeRFs need to trace a continuous path along the rays by selecting a discrete set of Gaussians that will be rendered along a specific ray.

\subsection{Optical Navigation for RPO}

Object detection using traditional computer vision techniques for optical navigation for RPO has long been used for autonomous and manned missions. However, more recently, CNN-based object detection techniques for relative RPO with cameras have garnered great interest due to its adaptability to unknown situations. Some well known implementations of this approach have been demonstrated \cite{aarestad_challenges_2020} to detect CubeSats with YOLOv3 on a Raspberry Pi (RPI) and a Neural Compute Stick 2 (NCS2). Others use Faster R-CNN to detect features \cite{chen_r-cnn-based_2020, viggh_training_2023}. The latter detects features of the Hubble space telescope by training the model on a synthetic dataset. Faraco et al. \cite{faraco2022instance} a Blender based tool called JINS to generate synthetic images of satellites and demonstrated segmentation and object detection performance of unknown satellites with Faster R-CNN and Mask R-CNN. Though Faster R-CNN shows high performance, it is computationally expensive to run in real-time on edge hardware. In these cases, the authors prefer accuracy over inference times.

This paper explores merging 3D reconstruction with object detection to improve predictions and identify pose of the spacecraft on achievable compute costs for on-board deployment. Mahendrakar et al. \cite{mahendrakar_real-time_2021, mahendrakar_use_2021, mahendrakar_performance_2022} used YOLOv5 models trained on images of satellites collected from the Internet and tested on HIL images under different lighting conditions. Further comparison studies were performed between YOLOv5 and Faster R-CNN and, YOLOv5 against YOLOv6, YOLOv7, YOLOv8 on Raspberry Pi \cite{Mahendrakar2024}. These studies showed high performance and concluded that YOLOv5 is best suited to be deployed on edge hardware for flight tests due to fast inference times as opposed to Faster R-CNN and, better performance on real-world satellite mock-up images as opposed to YOLOv6 through YOLOv8. The authors demonstrated \cite{mahendrakar_autonomous_2021, mahendrakar_autonomous_2023} HIL autonomous swarm satellite RPO tests with drones by merging object detection outputs of solar panels and body locations from a single observer with artificial potential field guidance algorithms. The drones collaboratively rendezvous to the attractive points while avoiding repulsive points on the detected features. Results from the flight tests revealed that the weakness of the autonomous RPO system is the accuracy of the predictions. They also state that using multiple observers in the system would improve the flight test results. However, this solution is impractical for real-world application to have that many satellites in close proximity. It is much preferred to deploy a single observer than multiple observers.

To streamline this solution, in \cite{caruso_3d_2023, nguyen2024} a single observer was used to perform R-bar and V-bar hops around the target spacecraft to capture images. These images were then used to perform 3D Reconstruction for geometry characterization of the unknown satellite. The authors investigated popular NeRF, D-NeRF, Instant NeRF, 3D Gaussian Splatting and 4D Gaussian Splatting to reconstruct the spacecraft model. The results concluded that 3D Gaussian Splatting shows the best run time and reconstruction performance.


\section{Methods}

\subsection{3D Gaussian Splatting}

The core idea behind 3D Gaussian Splatting \cite{kerbl_3d_2023} is to treat each point as a 3D Gaussian distribution with a mean equal to the point's position and a covariance matrix that determines the shape and orientation of the distribution. The Gaussian distribution is defined as:

\begin{equation}
G(\mathbf{x}) = e^{-\frac{1}{2}(\mathbf{x})^T \Sigma^{-1}(\mathbf{x})}
\end{equation}

where $\mathbf{x}$ is a 3D position vector relative to the point's center, and $\Sigma$ is the covariance matrix in the world space.

The covariance matrix $\Sigma$ is a symmetric, positive semi-definite matrix that encodes the shape and orientation of the Gaussian distribution. It can be decomposed into a rotation matrix $R$ and a scaling matrix $S$:

\begin{equation}
\Sigma = RSS^T R^T
\end{equation}

The rotation matrix $R$ determines the orientation of the Gaussian distribution, while the scaling matrix $S$ controls the extent of the distribution along each axis.

To render the point set, each Gaussian distribution is projected onto the screen using the camera's projection matrix. The projected Gaussian distribution is then rasterized, and its contribution is accumulated in the image buffer. The projection of the Gaussian distribution is performed by applying a transformation to the covariance matrix according to \citet{zwicker2001}:

\begin{equation}
\Sigma' = JW\Sigma W^T J^T
\end{equation}

where $J$ is the Jacobian matrix of the projection transformation, and $W$ is the world-to-camera transformation matrix.

The projected Gaussian distributions are then rasterized, and their contributions are accumulated in the image buffer. By accumulating the contributions of all the projected Gaussian distributions, the final image is formed, providing a smooth and continuous representation of the point set. The resulting image quality can be controlled by adjusting the size and shape of the Gaussian distributions, as well as the number of points in the set.

\subsection{Camera Generation}

Given a set of $n$ initial camera poses represented by their $4 \times 4$ transformation matrices $\mathbf{T}_i$, where $i \in {1, \ldots, n}$, we aim to generate novel camera viewpoints around a synthetic rendezvous path. Matrix $\mathbf{T}_i$ contains right ($\mathbf{r}_i$), up ($\mathbf{u}_i$), forward ($\mathbf{f}_i$), and camera position vectors ($\mathbf{c}_i$), each one is an $\mathbb{R}^3$ column vector, as demonstrated in \eqref{eq:T-matrix}.

\begin{equation}
    \mathbf{T}_i = \begin{bmatrix}
{r}_{xi} & {u}_{xi} & {f}_{xi} & {c}_{xi} \\
{r}_{yi} & {u}_{yi} & {f}_{yi} & {c}_{yi} \\
{r}_{zi} & {u}_{zi} & {f}_{zi} & {c}_{zi} \\
0 & 0 & 0 & 1 \\
\end{bmatrix}
\label{eq:T-matrix}
\end{equation}

\textbf{Nearest point of camera attention.} To generate new camera poses, we first estimate a reference point $\mathbf{p} \in \mathbb{R}^3$ that all initial cameras are roughly pointed towards (we'll call this the camera attention center). Inspired by \citet{Han2010}, we formulate this as a nearest point to multiple lines in 3D space least-squares problem:
\begin{align}
    \min_{\mathbf{p}} \lVert \mathbf{A}\mathbf{x} - \mathbf{C} \rVert_2^2,
\end{align}
where $\mathbf{C} \in \mathbb{R}^{3n \times 1}$ is a column vector that concatenates all camera positions $[\mathbf{c}_1^{\top}, \mathbf{c}_2^{\top}, \ldots, \mathbf{c}_n^{\top}]^{\top}$, $\mathbf{x} = [\mathbf{p}^{\top}, a_1, a_2, \ldots, a_n]^{\top}$ is the solution vector containing the estimated attention center point $\mathbf{p} \in \mathbb{R}^3$ in the first three elements and scalar values $a_i$ represent the parameters of these line equations:
\begin{align}
    \frac{x - c_{xi}}{f_{xi}} = \frac{y - c_{yi}}{f_{yi}} = \frac{z - c_{zi}}{f_{zi}} = a_i
\end{align}

Expanding this equation yields a point $\mathbf{p}_i$ on the line corresponding to camera $i$:
\begin{align}
    \mathbf{p}_i = \mathbf{f}_i \cdot a_i + \mathbf{c}_i
\end{align}
and $\mathbf{A} \in \mathbb{R}^{3n \times (n+3)}$ is a block matrix constructed as follows:
\begin{equation}
\mathbf{A} = \begin{bmatrix}
\mathbf{I}_3 & -\mathbf{f}_1 & \mathbf{0} & \cdots & \mathbf{0} \\
\mathbf{I}_3 & \mathbf{0} & -\mathbf{f}_2 & \cdots & \mathbf{0} \\
\vdots & \vdots & \vdots & \ddots & \vdots \\
\mathbf{I}_3 & \mathbf{0} & \mathbf{0} & \cdots & -\mathbf{f}_n
\end{bmatrix}
\end{equation}
where $\mathbf{I}_3$ represents the $3 \times 3$ identity matrix, $\mathbf{f}_i \in \mathbb{R}^{3 \times 1}$ is the column forward vector $[f_{xi}, f_{yi}, f_{zi}]^{\top}$ for each camera $i$, and $\mathbf{0}$ denotes a $3 \times 1$ column vector of zeros.

\begin{figure}[H]
    \centering
    \includegraphics[width=0.5\linewidth]{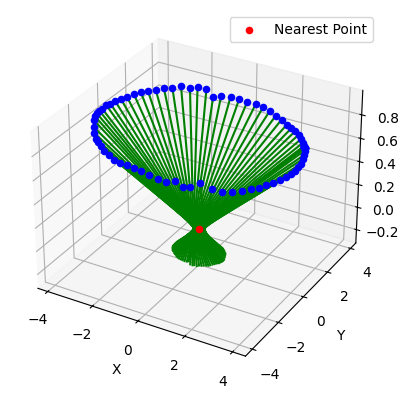}
    \caption{Nearest point to non-intersecting camera lines}
    \label{fig:nearest-point}
\end{figure}

The solution to this least-squares problem can be obtained using standard linear algebra techniques, such as the pseudoinverse or singular value decomposition (SVD), yielding the estimated camera attention center $\mathbf{p}$. Example of the solution is shown in Figure~\ref{fig:nearest-point}, where we applied the method to find the center of attention (red) from 84 non-intersecting camera lines, constructed from original camera positions (blue) and their forward vectors (green).

\begin{figure}[h]
    \centering
    \begin{subfigure}{0.4\textwidth}
        \centering
        \includegraphics[width=\linewidth]{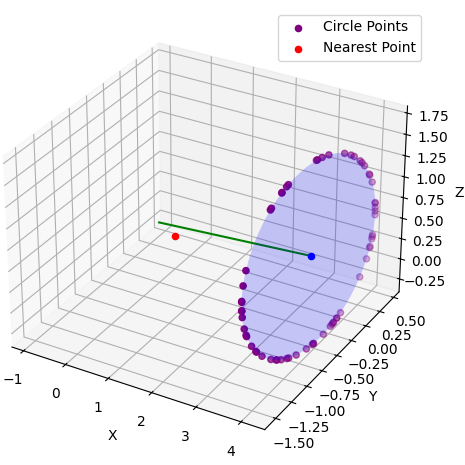}
        \caption{Circular, random}
        \label{fig:circle_rand}
    \end{subfigure}
    \begin{subfigure}{0.4\textwidth}
        \centering
        \includegraphics[width=\linewidth]{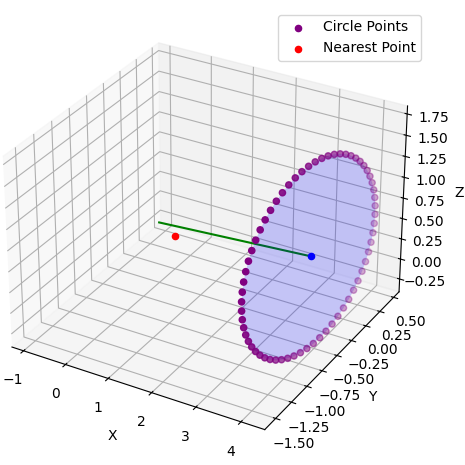}
        \caption{Circular, equidistant}
        \label{fig:circle_equi}
    \end{subfigure}
    
    \vspace{1em}
    
    \begin{subfigure}{0.4\textwidth}
        \centering
        \includegraphics[width=\linewidth]{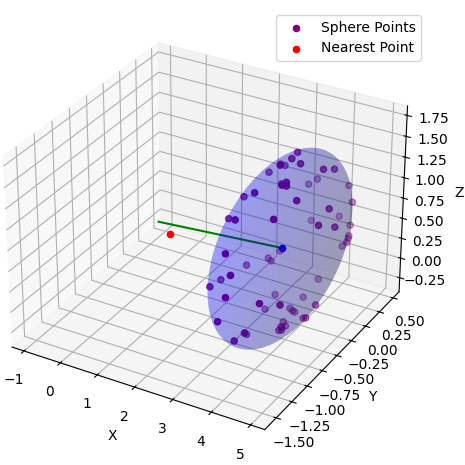}
        \caption{Spherical, random}
        \label{fig:sphere_rand}
    \end{subfigure}
    \begin{subfigure}{0.4\textwidth}
        \centering
        \includegraphics[width=\linewidth]{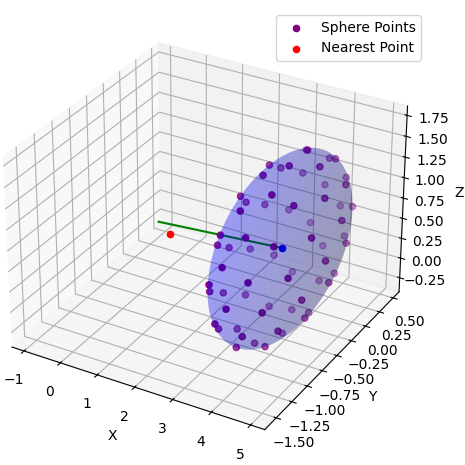}
        \caption{Spherical, semi-equidistant}
        \label{fig:sphere_fibb}
    \end{subfigure}
    
    \caption{Visualization for generation of 64 new cameras with radius 1 from the original camera position. Note that the sphere is deformed due to scaling difference between x, y, z axes}
    \label{fig:cam-gen}
\end{figure}

Two methods are provided for generating $m$ novel poses per initial camera, visualized in Figure~\ref{fig:cam-gen}:

\begin{itemize}
\item \textbf{Circular}: For each $\mathbf{c}_i$, generate poses $\mathbf{c}^\prime_{ij}$ on a circle of radius $r$ in the plane perpendicular to $\mathbf{v}_i = \mathbf{p} - \mathbf{c}_i$, centered at $\mathbf{c}_i$. An orthonormal basis ${\mathbf{u}_i, \mathbf{w}_i}$ for this plane is constructed using cross products with $\mathbf{v}_i$. The new positions are parameterized as:
$$\mathbf{c}^\prime_{ij} = \mathbf{c}_i + r\cos(\theta_j)\mathbf{u}_i + r\sin(\theta_j)\mathbf{w}_i, \quad j \in {1, \ldots, m}$$
where $\theta_j$ are equispaced or randomly generated angles in $[0, 2\pi]$.

\item \textbf{Spherical}: For each $\mathbf{c}_i$, generate poses $\mathbf{c}^\prime_{ij}$ on a sphere of radius $r$ centered at $\mathbf{c}_i$. The positions are parameterized using spherical coordinates for $j=1,...,m$:
$$\mathbf{c}_{ij} = \mathbf{c}_i + r\begin{bmatrix}
\sin(\phi_j)\cos(\theta_j) \\
\sin(\phi_j)\sin(\theta_j) \\
\cos(\phi_j)
\end{bmatrix}$$
where $\phi_j$ and $\theta_j$ are sampled either randomly or using a Fibonacci lattice \cite{gonzalez2010} for semi-uniform coverage.
\end{itemize}

The forward vectors $\mathbf{f}^\prime_{ij}$ for the new cameras are computed as unit vectors pointing towards $\mathbf{p}$. Finally, the 4x4 transformation matrices $\mathbf{T}^\prime_{ij}$ for the novel poses are constructed by augmenting $\mathbf{c}^\prime_{ij}$, $\mathbf{f}^\prime_{ij}$, and reusing the original camera axes, similar to \eqref{eq:T-matrix}.

\subsection{YOLOv5 Ensembling}

YOLOv5 \cite{jocher_yolov5_2020} is a single-stage object detector selected for use in this research. It has shown the most success in generalizing to perform component detection on unknown satellites \cite{Mahendrakar2024}. Further, we choose the ``small'' architecture, as it can run at sufficient framerates on current spaceflight hardware. YOLO performs uses a single CNN to map input image pixels directly to bounding box and class predictions.

The images are input to the YOLO architecture, which consists of a backbone CNN that extracts visual features, a neck that synthesizes those features, and a head that uses these features to predict the bounding boxes. This head makes a coarse regular grid on the input image, estimates the probability an object is centered in each grid square, estimates class probabilities of potential objects in each grid square, and generates a fixed number of bounding box coordinates and dimensions centered in each grid square. This results in many redundant boxes, of which the least confident of overlapping boxes are tossed out using non-maximum suppression (NMS).

We implement YOLOv5 using the public GitHub repository \cite{jocher_yolov5_2020}. Weights trained for satellite component detection in \cite{mahendrakar_autonomous_2023} is used. The model was trained using the Adam optimizer \cite{kingma_adam_2017} on the Web Satellite Dataset (WSD) \cite{Mahendrakar2024}. We test on a real-life mock-up of a satellite in two cases:
\begin{itemize}
    \item \textbf{Known satellite}: the object detector has seen the mock-up during training (images do not overlap with training images)
    \item \textbf{Unknown satellite}: the object detector has never seen the mock-up during training
\end{itemize}
This allows us to assess our component detection performance in each case.

Prior research \cite{mahendrakar_real-time_2021, dung_spacecraft_2021, mahendrakar_performance_2022} has demonstrated CNN-based object detection is effective but imperfect at detecting components of satellites, especially unknown satellites. Hence, we propose a novel ensembling technique to improve performance. The technique is as follows:
\begin{enumerate}
    \item Capture an image of a satellite from an observer at a known viewing position. 
    \item Render additional images from observer positions surrounding vantage points using 3D Gaussian Splatting. 
    \item Process ground truth and 3D-rendered images with YOLOv5 \cite{jocher_yolov5_2020} to detect components of satellites in each viewframe.
    \item Ensemble detections from real image and synthetic renders to enhance real-image detections.
\end{enumerate}

The synthetic views and inferences made on them will create an ensemble of predictions. These predictions will be averaged and used to add confidence to true positive or true negative predictions from the real viewing angle or to override errors. Ultimately, this results in more accurate and reliable detection of spacecraft components.

\subsection{Datasets}
Due to the nature of this work, we use two different datasets. The object detection dataset primarily consists of images from the internet and the 3DGS dataset consists of HIL images of a mockup satellite.

The object detection training dataset consists of 1231 images of satellites, same as the one used in \cite{Mahendrakar2024}. In each image, if present, solar panels, antennas, thruster and body are annotated. The dataset is augmented to add random cutout boxes, Gaussian blur, salt pepper noise, shear and rotation prior to training and testing. A summary of the object detection dataset containing number of annotations is shown in the figure below. 

\begin{figure}[H]
    \centering
    \includegraphics[width=0.6\linewidth]{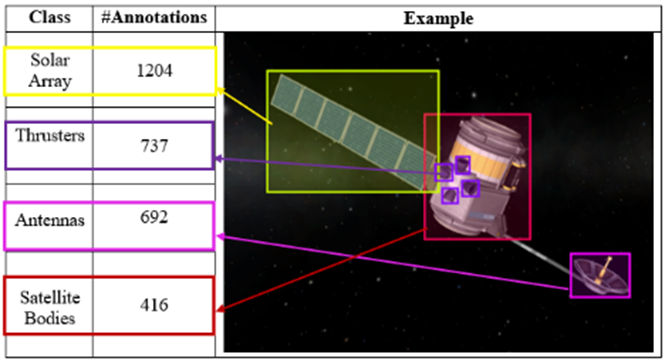}
    \captionsetup{justification=centering}
    \caption{Object Detection Dataset \cite{Mahendrakar2024}}
    \label{fig:Object Detection Dataset}
\end{figure}

\begin{figure}[H]
    \centering
    \includegraphics[width=0.6\linewidth]{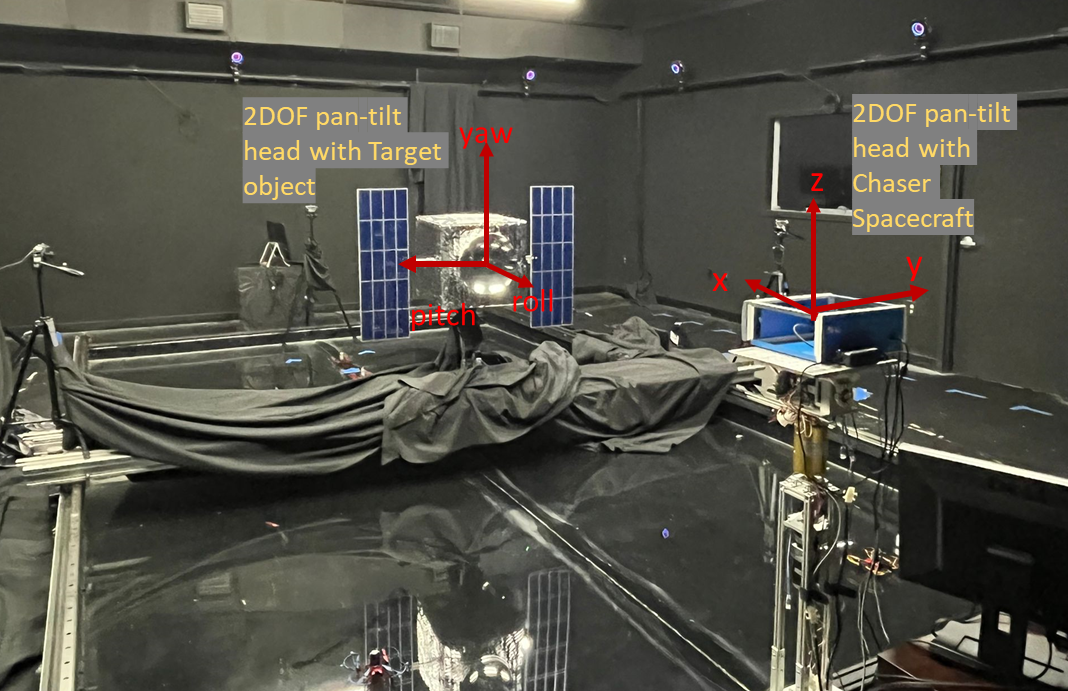}
    \captionsetup{justification=centering}
    \caption{ORION Testbed \cite{wilde_orion_2016}}
    \label{fig:ORION}
\end{figure}

The testing dataset for object detection and the 3DGS datasets are the same. They were captured on the ORION testbed \cite{wilde_orion_2016}. The ORION testbed consists of a planar Cartesian maneuver kinematics simulator shown in Figure~\ref{fig:ORION}. It has 5.5m x 3.5m workspace. The simulator is also equipped with a 2 DOF motion table that can translate up to 0.25 m/s and accelerate up to $1 m/s^2$. It also offers two custom designed pan-tilt mechanisms that can yaw infinitely and pitch between $+/-90^\circ$. The target and the chaser spacecraft can be interchanged as needed. The target can be modified to add different shaped components such as solar panels and antennas. Lighting conditions are simulated by a Hilio D12 LED panel with a color temperature of 5600 K (daylight-balanced) rated for 350 W of power with adjustable intensity.       

In the images used for this study, the mock-up satellite yaws at 10$^\circ$/s and the chaser satellite captures images positioned 5 ft away, simulating V-bar station keeping maneuver. Images are captured at 5$^\circ$ increments. To denoise the background we used a green screen behind the mock-up and performed chroma key compositing to replace green pixels with black. We set the lighting to 100\% to simulate worst case conditions. The results of the test setup after chroma key compositing is shown below.

\begin{figure}[H]
    \centering
    \begin{subfigure}{0.45\textwidth}
        \centering
        \includegraphics[width=1\linewidth]{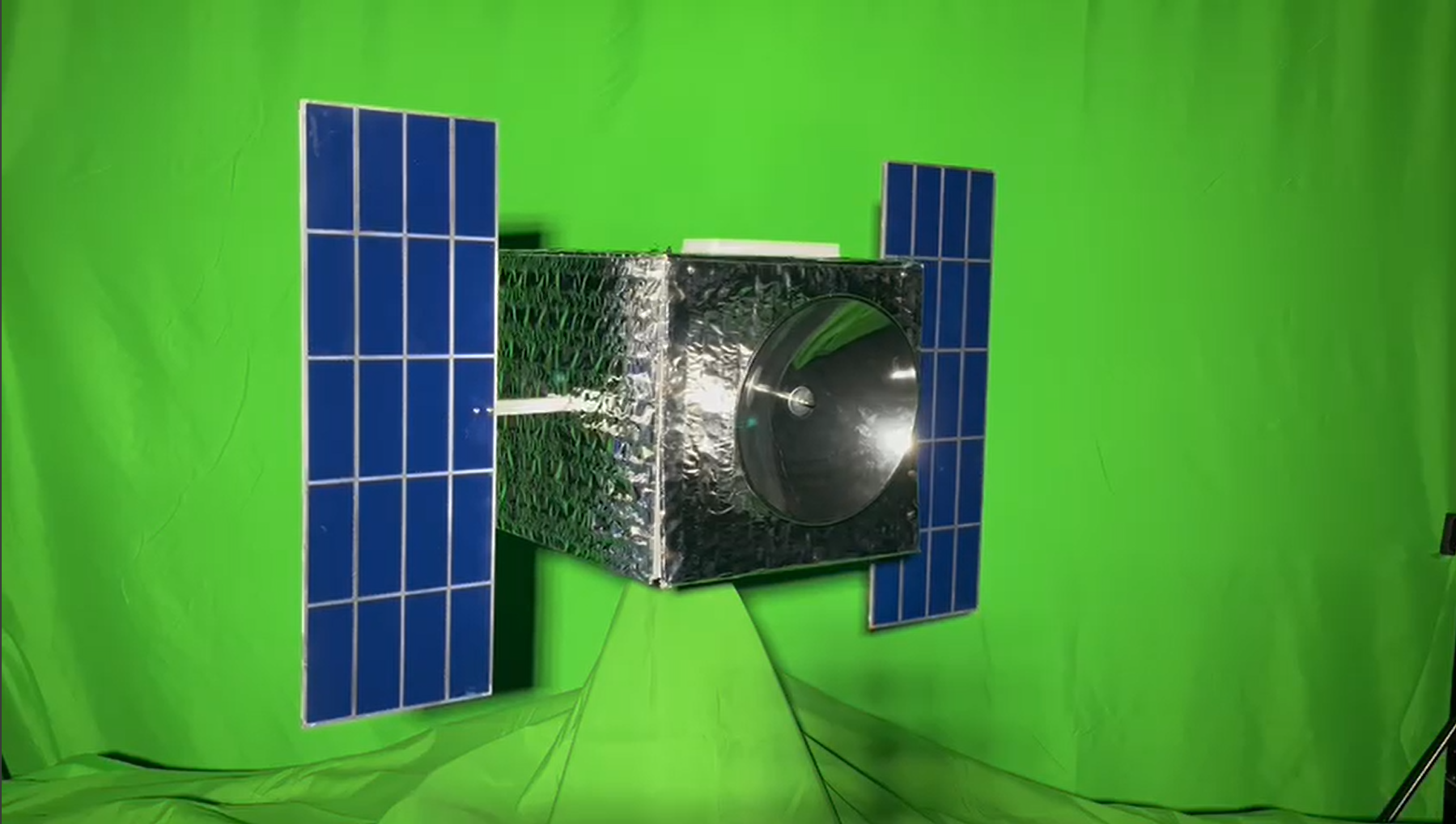}
        \caption{Original Image \cite{caruso_3d_2023, nguyen2024}}
    \end{subfigure}
    \begin{subfigure}{0.45\textwidth}
        \centering
        \includegraphics[width=\linewidth]{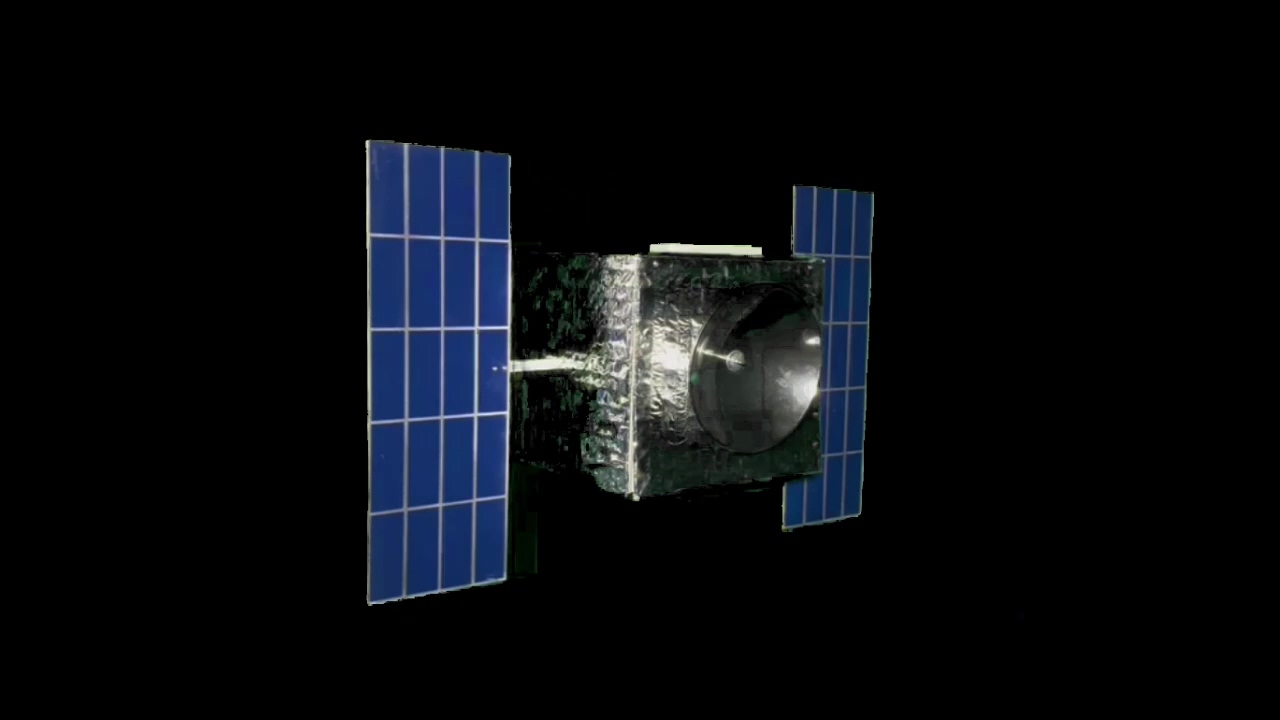}
        \caption{After Chroma Keying  \cite{caruso_3d_2023, nguyen2024}}
    \end{subfigure}
    \captionsetup{justification=centering}
\end{figure}

\subsection{Performance Evaluation}

Object detection performance will be measured in terms of both classification and localization accuracy. We use the standard mean average precision (mAP) metrics, mAP@0.5 and mAP@0.5:0.95, as 1-number summaries of object detection performance.

The quality of the 3D model produced by 3DGS is assessed using the structural similarity index (SSIM), peak signal-to-noise ratio (PSNR), and learned perceptual image patch similarity (LPIPS) \cite{zhang2018unreasonable}.

\section{Results}

\begin{figure}[H]
    \centering
    \begin{subfigure}{0.33\linewidth}
        \centering
        \includegraphics[width=\linewidth]{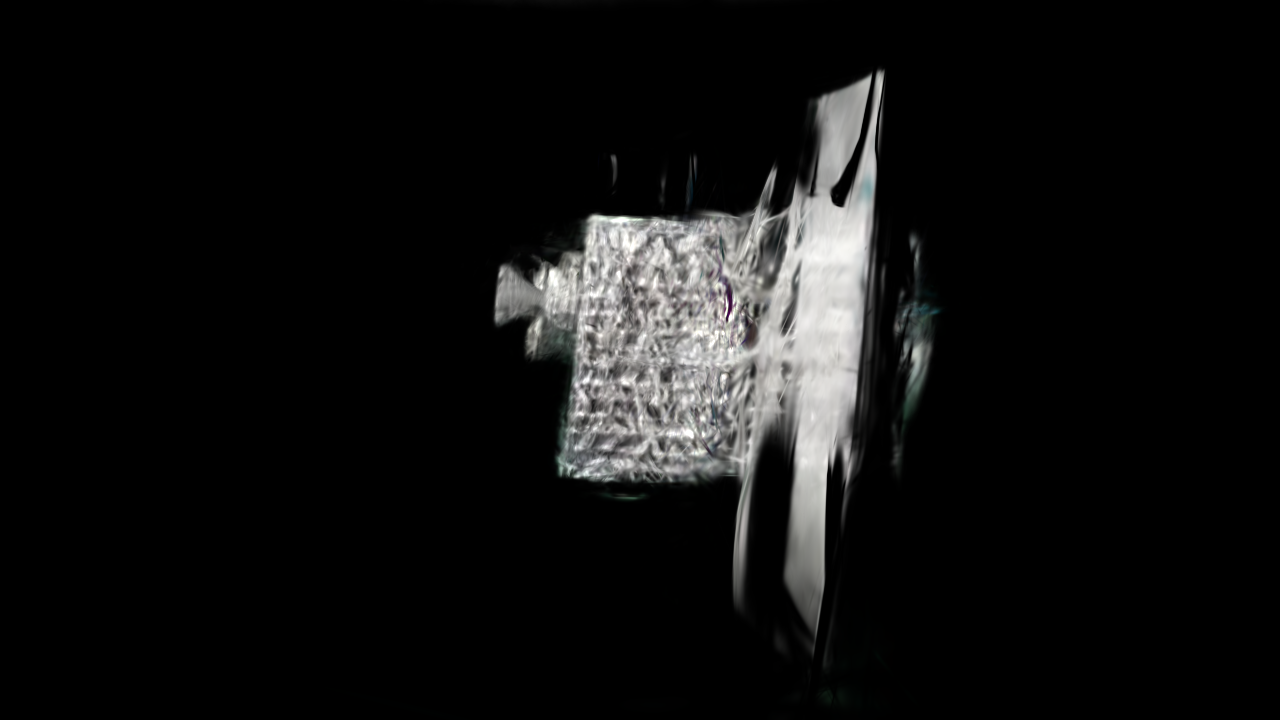}
        \caption{Generated camera render \#19}
    \end{subfigure}
    \hfill
    \begin{subfigure}{0.33\linewidth}
        \centering
        \includegraphics[width=\linewidth]{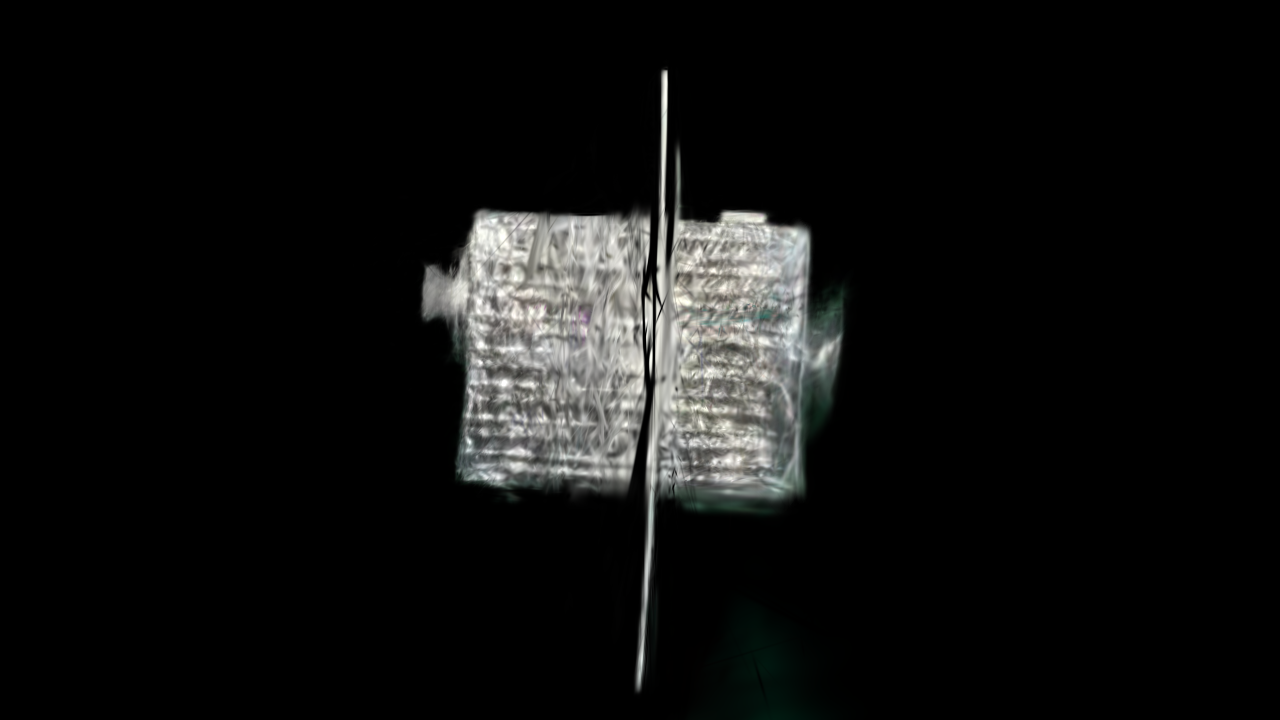}
        \caption{True camera render}
    \end{subfigure}    
    \hfill
    \begin{subfigure}{0.33\linewidth}
        \centering
        \includegraphics[width=\linewidth]{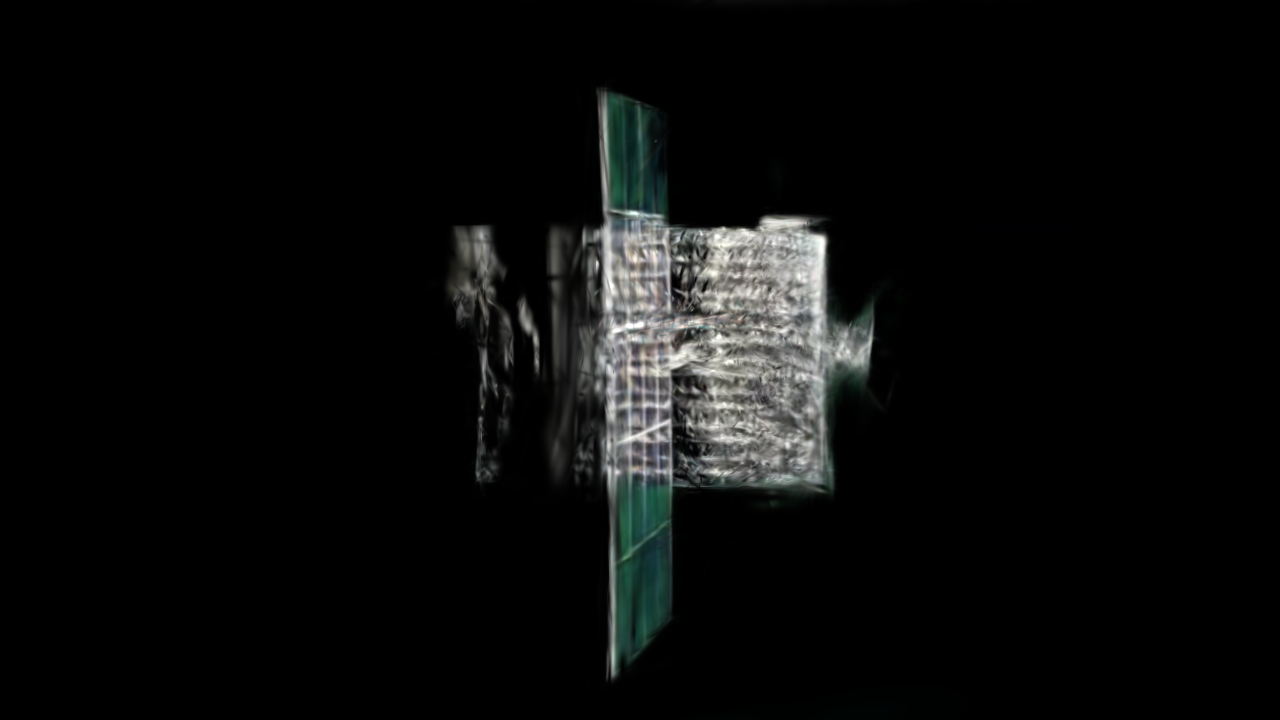}
        \caption{Generated camera render \#50}
    \end{subfigure}    
    \caption{High quality render of true camera and 2 of the generated camera array from the same true camera}\label{fig:cam-gen-render}
\end{figure}

3DGS is trained to produce a 3D representation of the scene by the approach of Nguyen et al. \cite{nguyen2024} The resulting quality of that work is replicated, achieving 0.9213 SSIM, 25.52 PSNR, and 0.0796 LPIPS. This is a high-quality model as seen in the renders it produces in Figure~\ref{fig:cam-gen-render}.

The SatSplatYOLO ensembling technique is customized to the nature of the satellite component detection task and the quality of the original object detector for this task. For each of the 84 ground truth camera views, we have 64 renders. YOLOv5 inference is run on both the ground truth frame and all 64 renders in each case to predict bounding boxes and their corresponding classifications (solar, body, antenna, or thruster) and confidence score. The goal of the ensembling is to exploit this batch of detections and enhance the original prediction.

For the detections from all 64 renders, we assign detections with bounding boxes with high mutual intersection over union (IoU) over a threshold to groups. For groups with sufficiently many detections, we find their majority class prediction and purity (i.e. the fraction of classifications in agreement with the majority). Groups with purity above a threshold are retained, and we merge redundant groups with the same classification with IoU over a threshold into one group. This gives several detection groups with high agreement on class prediction and many bounding boxes clustered together. The four thresholds here are grid searched for best performance in each case in the table below.

\begin{figure}
    \centering
    \includegraphics[width=\textwidth, height=4.5cm]{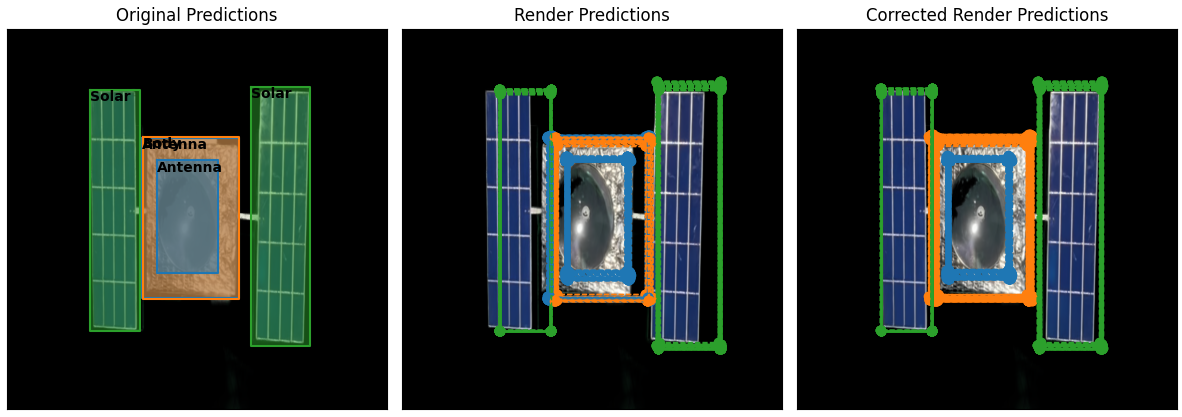}
    \caption{YOLOv5 inferences on the original  image (left), the 3DGS renders (center), and final SatSplatYOLO predictions (right).}
    \label{fig:ensembles}
\end{figure}

These clusters experience an consistent translation offset due to the approximation of the nearest point algorithm as seen in Figure~\ref{fig:ensembles} (center). We then perform a translation correction by mapping the mean bounding box from the detection groups to high-confidence YOLOv5 predictions on the original images. The base YOLOv5 model makes nearly no mistakes in detecting satellite bodies, regardless of whether the target is a known or an unknown satellite, so it is our typical anchor. In addition, we make no change to body predictions.

Finally, we make our corrected predictions. If a render detection group coincides in location with an original detection, we make its confidence $\max(\text{conf, purity})$, replace the bounding box with a weighted average (by confidence and purity) of the original bounding box and mean bounding box of the group, and assign the group-predicted class. If a render detection group does not coincide with the location of an original detection, we make a new prediction to correct original model's likely false negative. This results further in removing ground truth predictions that do not coincide with a render detection group, suppressing false positives.

Results are shown in Table~\ref{tab:yolov5_results} for many experimental cases. The left half of the table is for a known satellite (some images of the satellite mock-up are used to help train the model) and the right half is for an unknown satellite (the model has never seen the mock-up). In addition, we try all four techniques discussed above for virtual camera positioning for rendering, each with radius values. We document the precision, recall, mAP@0.5 for each object class and model. SatSplatYOLO improves results for unknown satellites significantly, but performance on known satellites is better without SatSplatYOLO.

\begin{table*}
\centering
\begin{tabular}{lccccccc}
\toprule[1.5pt]
 & & \multicolumn{3}{c}{Known Satellite} & \multicolumn{3}{c}{Unknown Satellite} \\
\cmidrule(lr){3-5} \cmidrule(lr){6-8}
 Cameras & Radius & Precision & Recall & mAP@0.5 & Precision & Recall & mAP@0.5 \\
\midrule[1.2pt]

Circular
 & 0.1 & 0.677 & \underline{0.710} & 0.692 & 0.444 & \textbf{0.460} & 0.402 \\
 (random) & 0.5 & 0.672 & 0.702 & 0.651 & 0.502 & 0.386 & \underline{0.440} \\
 & 1.0 & 0.577 & 0.597 & 0.548 & 0.392 & 0.336 & 0.361
 \\
\midrule

Circular 
 & 0.1 & 0.720 & 0.690 & 0.696 & 0.440 & 0.412 & 0.404
 \\
 (equidistant) & 0.5 & 0.728 & 0.704 & \underline{0.701} & 0.455 & 0.400 & 0.421
\\
 & 1.0 & 0.633 & 0.596 & 0.603 & 0.435 & 0.339 & 0.386
 \\
\midrule
Spherical & 1.0 & 0.710 & 0.709 & \underline{0.701} & 0.450 & 0.407 & 0.404
 \\
 (random) & 0.5 & 0.627 & 0.700 & 0.626 & 0.480 & 0.390 & 0.430
 \\
 & 1.0 & 0.579 & 0.657 & 0.554 & 0.550 & 0.332 & 0.429
 \\
\midrule
Spherical
 & 0.1 & 0.695 & 0.694 & 0.692 & 0.469 & \underline{0.417} & 0.414
 \\
 (semi-equidistant) & 0.5 & 0.711 & 0.688 & 0.684 & 0.475 & 0.400 & 0.433
 \\
 & 1.0 & \underline{0.740} & 0.604 & 0.652 & \textbf{0.590} & 0.384 & \textbf{0.485}
 \\ \midrule
 Ground Truth & N/A & \textbf{0.770} & \textbf{0.728} & \textbf{0.758} & \underline{0.544} & 0.384 & 0.432 \\
\bottomrule[1.5pt]
\end{tabular}
\caption{SatSplatYOLO Results for Different Camera Paths. \textbf{Bolded} results are best metrics for each class across 13 different camera configurations and radius combinations. \underline{Underlined} results are second best.}
\label{tab:yolov5_results}
\end{table*}

In Figure~\ref{fig:ensembles}, we see an example of the ground truth image-based predictions, the rendered image predictions, and the final predictions corrected for translation errors in the renders and class prediction averaging. In the original image (left), we see correct detections of the two solar panels, antenna, and body, but there is also a false detection of an antenna nearly fully overlapping the body. In the rendered images, we see the same 5 predictions, where some the detections that \textit{should} correspond to the body are classified as body in some renders and antennas in others. In addition, the imperfect viewing angle coming from SfM and the nearest point algorithm result in predictable translation errors in localization. The SatSplatYOLO model combines the original and render predictions to correct the translation error and then suppresses the incorrect false antenna predictions as a minority class of the ensemble of predictions around the body.

\section{Conclusion}

In this work, we have proposed a method SatSplatYOLO for detecting components of satellites by learning a 3D representation of the model, rendering novel views of the satellite, and ensembling YOLOv5 (small) over it. The computational costs are sufficiently low to feasibly run on near-term spaceflight hardware, based on image data captured through one inspection orbit.

This method not only improves prediction accuracy but also provides a first step to reliable pose estimation of unknown non-cooperative satellites. As long as at least three non-symmetrical objects are detected accurately (per our test satellite mock-up) during any phase of the mission, the pose of the target RSO can be estimated from the initial 3D reconstruction. Estimating the pose of the spacecraft this way provides more confidence than most methods in the literature as it provides a holistic understanding of the scenario. 


\bibliographystyle{ieeetr}  
\bibliography{main}

\begin{thebibliography}{10}

\bibitem{szegedy_deep_2013}
Christian Szegedy, Alexander Toshev, and Dumitru Erhan.
\newblock Deep {Neural} {Networks} for {Object} {Detection}.
\newblock In {\em Advances in {Neural} {Information} {Processing} {Systems}}, volume~26. Curran Associates, Inc., 2013.

\bibitem{sermanet_overfeat_2014}
Pierre Sermanet, David Eigen, Xiang Zhang, Michael Mathieu, Rob Fergus, and Yann LeCun.
\newblock {OverFeat}: {Integrated} {Recognition}, {Localization} and {Detection} using {Convolutional} {Networks}, February 2014.
\newblock arXiv:1312.6229 [cs].

\bibitem{noauthor_papers_nodate}
Papers with {Code} - {Object} {Detection}.

\bibitem{noauthor_papers_nodate-1}
Papers with {Code} - {Real}-{Time} {Object} {Detection}.

\bibitem{he_spatial_2015}
Kaiming He, X.~Zhang, Shaoqing Ren, and J.~Sun.
\newblock Spatial {Pyramid} {Pooling} in {Deep} {Convolutional} {Networks} for {Visual} {Recognition}.
\newblock {\em IEEE Transactions on Pattern Analysis and Machine Intelligence}, 37(9):1904--1916, 2015.

\bibitem{he_deep_2016}
Kaiming He, Xiangyu Zhang, Shaoqing Ren, and Jian Sun.
\newblock Deep {Residual} {Learning} for {Image} {Recognition}.
\newblock pages 770--778, 2016.

\bibitem{lin_feature_2017}
Tsung-Yi Lin, Piotr Dollar, Ross Girshick, Kaiming He, Bharath Hariharan, and Serge Belongie.
\newblock Feature {Pyramid} {Networks} for {Object} {Detection}.
\newblock pages 2117--2125, 2017.

\bibitem{liu_path_2018}
Shu Liu, Lu~Qi, Haifang Qin, Jianping Shi, and Jiaya Jia.
\newblock Path {Aggregation} {Network} for {Instance} {Segmentation}, September 2018.
\newblock arXiv:1803.01534 [cs].

\bibitem{wang_cspnet_2020}
Chien-Yao Wang, Hong-Yuan~Mark Liao, Yueh-Hua Wu, Ping-Yang Chen, Jun-Wei Hsieh, and I.-Hau Yeh.
\newblock {CSPNet}: {A} {New} {Backbone} {That} {Can} {Enhance} {Learning} {Capability} of {CNN}.
\newblock pages 390--391, 2020.

\bibitem{krizhevsky_imagenet_2012}
Alex Krizhevsky, Ilya Sutskever, and Geoffrey~E Hinton.
\newblock {ImageNet} {Classification} with {Deep} {Convolutional} {Neural} {Networks}.
\newblock In F.~Pereira, C.~J. Burges, L.~Bottou, and K.~Q. Weinberger, editors, {\em Advances in {Neural} {Information} {Processing} {Systems}}, volume~25. Curran Associates, Inc., 2012.

\bibitem{liu_ssd_2016}
Wei Liu, Dragomir Anguelov, Dumitru Erhan, Christian Szegedy, Scott Reed, Cheng-Yang Fu, and Alexander~C. Berg.
\newblock {SSD}: {Single} {Shot} {MultiBox} {Detector}.
\newblock In Bastian Leibe, Jiri Matas, Nicu Sebe, and Max Welling, editors, {\em Computer {Vision} – {ECCV} 2016}, Lecture {Notes} in {Computer} {Science}, pages 21--37, Cham, 2016. Springer International Publishing.

\bibitem{redmon_you_2016}
Joseph Redmon, Santosh Divvala, Ross Girshick, and Ali Farhadi.
\newblock You {Only} {Look} {Once}: {Unified}, {Real}-{Time} {Object} {Detection}.
\newblock pages 779--788, 2016.

\bibitem{bochkovskiy_yolov4_2020}
Alexey Bochkovskiy, Chien-Yao Wang, and Hong-Yuan~Mark Liao.
\newblock {YOLOv4}: {Optimal} {Speed} and {Accuracy} of {Object} {Detection}, April 2020.
\newblock arXiv:2004.10934 [cs, eess].

\bibitem{jocher_yolov5_2020}
Glenn Jocher.
\newblock {YOLOv5} by {Ultralytics}, May 2020.

\bibitem{li_yolov6_2022}
Chuyi Li, Lulu Li, Hongliang Jiang, Kaiheng Weng, Yifei Geng, Liang Li, Zaidan Ke, Qingyuan Li, Meng Cheng, Weiqiang Nie, Yiduo Li, Bo~Zhang, Yufei Liang, Linyuan Zhou, Xiaoming Xu, Xiangxiang Chu, Xiaoming Wei, and Xiaolin Wei.
\newblock {YOLOv6}: {A} {Single}-{Stage} {Object} {Detection} {Framework} for {Industrial} {Applications}, September 2022.
\newblock arXiv:2209.02976 [cs].

\bibitem{wang_yolov7_2023}
Chien-Yao Wang, Alexey Bochkovskiy, and Hong-Yuan~Mark Liao.
\newblock {YOLOv7}: {Trainable} {Bag}-of-{Freebies} {Sets} {New} {State}-of-the-{Art} for {Real}-{Time} {Object} {Detectors}.
\newblock pages 7464--7475, 2023.

\bibitem{jocher_yolov8_2023}
Glenn Jocher, Ayush Chaurasia, and Jing Qiu.
\newblock {YOLOv8} by {Ultralytics}, January 2023.

\bibitem{mildenhall_nerf_2021}
Ben Mildenhall, Pratul~P. Srinivasan, Matthew Tancik, Jonathan~T. Barron, Ravi Ramamoorthi, and Ren Ng.
\newblock {NeRF}: representing scenes as neural radiance fields for view synthesis.
\newblock {\em Communications of the ACM}, 65(1):99--106, December 2021.

\bibitem{muller_instant_2022}
Thomas Müller, Alex Evans, Christoph Schied, and Alexander Keller.
\newblock Instant {Neural} {Graphics} {Primitives} with a {Multiresolution} {Hash} {Encoding}.
\newblock {\em ACM Trans. Graph.}, 41(4):102:1--102:15, July 2022.
\newblock Place: New York, NY, USA Publisher: ACM.

\bibitem{kerbl_3d_2023}
Bernhard Kerbl, Georgios Kopanas, Thomas Leimkuehler, and George Drettakis.
\newblock {3D} {Gaussian} {Splatting} for {Real}-{Time} {Radiance} {Field} {Rendering}.
\newblock {\em ACM Transactions on Graphics}, 42(4):139:1--139:14, July 2023.

\bibitem{schonberger_structure--motion_2016}
Johannes~L. Schonberger and Jan-Michael Frahm.
\newblock Structure-{From}-{Motion} {Revisited}.
\newblock pages 4104--4113, 2016.

\bibitem{aarestad_challenges_2020}
Jim Aarestad, Andrew Cochrane, Matthew Hannon, Evan Kain, Craig Kief, Steven Lindsley, and Brian Zufelt.
\newblock Challenges and {Opportunities} for {CubeSat} {Detection} for {Space} {Situational} {Awareness} using a {CNN}.
\newblock {\em Small Satellite Conference}, August 2020.

\bibitem{chen_r-cnn-based_2020}
Yulang Chen, Jingmin Gao, and Kebei Zhang.
\newblock R-{CNN}-{Based} {Satellite} {Components} {Detection} in {Optical} {Images}.
\newblock {\em International Journal of Aerospace Engineering}, 2020:e8816187, October 2020.
\newblock Publisher: Hindawi.

\bibitem{viggh_training_2023}
Herbert Viggh, Sean Loughran, Yaron Rachlin, Ross Allen, and Jessica Ruprecht.
\newblock Training {Deep} {Learning} {Spacecraft} {Component} {Detection} {Algorithms} {Using} {Synthetic} {Image} {Data}.
\newblock In {\em 2023 {IEEE} {Aerospace} {Conference}}, pages 1--13, March 2023.
\newblock ISSN: 1095-323X.

\bibitem{faraco2022instance}
Niccol{\`o} Faraco, Michele Maestrini, and Pierluigi Di~Lizia.
\newblock Instance segmentation for feature recognition on noncooperative resident space objects.
\newblock {\em Journal of Spacecraft and Rockets}, 59(6):2160--2174, 2022.

\bibitem{mahendrakar_real-time_2021}
Trupti Mahendrakar, Ryan~T. White, Markus Wilde, Brian Kish, and Isaac Silver.
\newblock Real-time {Satellite} {Component} {Recognition} with {YOLO}-{V5}.
\newblock In {\em 35th {Annual} {Small} {Satellite} {Conference}}, July 2021.

\bibitem{mahendrakar_use_2021}
Trupti Mahendrakar, Markus Wilde, and Ryan~T. White.
\newblock Use of {Artificial} {Intelligence} for {Feature} {Recognition} and {Flightpath} {Planning} {Around} {Non}-{Cooperative} {Resident} {Space} {Objects}.
\newblock In {\em {ASCEND} 2021}. American Institute of Aeronautics and Astronautics, 2021.
\newblock \_eprint: https://arc.aiaa.org/doi/pdf/10.2514/6.2021-4123.

\bibitem{mahendrakar_performance_2022}
Trupti Mahendrakar, Andrew Ekblad, Nathan Fischer, Ryan~T. White, Markus Wilde, Brian Kish, and Isaac Silver.
\newblock Performance {Study} of {YOLOv5} and {Faster} {R}-{CNN} for {Autonomous} {Navigation} around {Non}-{Cooperative} {Targets}.
\newblock In {\em 2022 {IEEE} {Aerospace} {Conference} ({AERO})}, pages 1--12. IEEE, 2022.

\bibitem{Mahendrakar2024}
Trupti Mahendrakar, Ryan~T. White, Madhur Tiwari, and Markus Wilde.
\newblock Real-time, on-board spacecraft feature detection with lightweight cnns.
\newblock {\em Submitted to AIAA Journal of Aerospace Information Systems}, 2024.

\bibitem{mahendrakar_autonomous_2021}
Trupti Mahendrakar.
\newblock {\em Autonomous {Feature} {Detection} for {Capture} {Path} {Planning} for {Rendezvous} and {Docking} with {Non}-{Cooperative} {Spacecraft}}.
\newblock PhD thesis, Florida Institute of Technology, December 2021.

\bibitem{mahendrakar_autonomous_2023}
Trupti Mahendrakar, Steven Holmberg, Andrew Ekblad, Emma Conti, Ryan~T. White, Markus Wilde, and Isaac Silver.
\newblock Autonomous {Rendezvous} with {Non}-cooperative {Target} {Objects} with {Swarm} {Chasers} and {Observers}.
\newblock In {\em 2023 {AAS}/{AIAA} {Spaceflight} {Mechanics} {Meeting}}. AAS/AIAA, January 2023.

\bibitem{caruso_3d_2023}
Basilio Caruso, Trupti Mahendrakar, Van~Minh Nguyen, Ryan~T. White, and Todd Steffen.
\newblock {3D} {Reconstruction} of {Non}-cooperative {Resident} {Space} {Objects} using {Instant} {NGP}-accelerated {NeRF} and {D}-{NeRF}, June 2023.
\newblock arXiv:2301.09060 [cs].

\bibitem{nguyen2024}
Van~Minh Nguyen, Emma Sandidge, Trupti Mahendrakar, and Ryan~T. White.
\newblock Characterizing satellite geometry via accelerated 3d gaussian splatting.
\newblock {\em Aerospace}, 11(3), 2024.

\bibitem{zwicker2001}
Matthias Zwicker, Hanspeter Pfister, Jeroen van Baar, and Markus Gross.
\newblock Surface splatting.
\newblock In {\em Proceedings of the 28th Annual Conference on Computer Graphics and Interactive Techniques}, SIGGRAPH '01, page 371–378, New York, NY, USA, 2001. Association for Computing Machinery.

\bibitem{Han2010}
L.~Han and J.~C. Bancroft.
\newblock {Nearest approaches to multiple lines in n-dimensional space}.
\newblock {\em {CREWES Research Report}}, 22:32.1--32.17, 2010.

\bibitem{gonzalez2010}
{\'A}lvaro Gonz{\'a}lez.
\newblock Measurement of areas on a sphere using fibonacci and latitude--longitude lattices.
\newblock {\em Mathematical Geosciences}, 42(1):49--64, Jan 2010.

\bibitem{kingma_adam_2017}
Diederik~P. Kingma and Jimmy Ba.
\newblock Adam: {A} {Method} for {Stochastic} {Optimization}, January 2017.
\newblock arXiv:1412.6980 [cs].

\bibitem{dung_spacecraft_2021}
Hoang~Anh Dung, Bo~Chen, and Tat-Jun Chin.
\newblock A {Spacecraft} {Dataset} for {Detection}, {Segmentation} and {Parts} {Recognition}.
\newblock In {\em 2021 {IEEE}/{CVF} {Conference} on {Computer} {Vision} and {Pattern} {Recognition} {Workshops} ({CVPRW})}, pages 2012--2019, Nashville, TN, USA, June 2021. IEEE.

\bibitem{wilde_orion_2016}
Markus Wilde, Brian Kaplinger, Tiauw Go, Hector Gutierrez, and Daniel Kirk.
\newblock {ORION}: {A} simulation environment for spacecraft formation flight, capture, and orbital robotics.
\newblock In {\em 2016 {IEEE} {Aerospace} {Conference}}, pages 1--14, March 2016.

\bibitem{zhang2018unreasonable}
Richard Zhang, Phillip Isola, Alexei~A Efros, Eli Shechtman, and Oliver Wang.
\newblock The unreasonable effectiveness of deep features as a perceptual metric.
\newblock In {\em Proceedings of the IEEE conference on computer vision and pattern recognition}, pages 586--595, 2018.

\end{thebibliography}

\end{document}